\def\year{2018}\relax
\documentclass[letterpaper]{article} 
\usepackage{aaai18}  
\usepackage{times}  
\usepackage{helvet}  
\usepackage{courier}  
\usepackage{url}  
\usepackage{graphicx}  
\usepackage{amsmath}
\usepackage{amsfonts}
\usepackage{amssymb}
\usepackage[plain, boxed]{algorithm}
\usepackage{algpseudocode}
\usepackage{multirow}
\usepackage{color}

\newcommand{\method}{{GAM}}
\newcommand{\methodmem}{{GAM-mem}}

\begin{document}
\title{Deep Graph Attention Model}


\author{
	John Boaz Lee\\
	Dept. of Computer Science\\
	Worcester Polytechnic Institute, MA
	\And
	Ryan Rossi\\
	Interactions and Analytics Lab\\
	Palo Alto Research Center, CA
	\And
	Xiangnan Kong\\
	Dept. of Computer Science\\
	Worcester Polytechnic Institute, MA
}

\maketitle
\begin{abstract}
Graph classification is a problem with practical applications in many different domains. Most of the existing methods take the entire graph into account when calculating graph features. In a graphlet-based approach, for instance, the entire graph is processed to get the total count of different graphlets or subgraphs. In the real-world, however, graphs can be both large and noisy with discriminative patterns confined to certain regions in the graph only. In this work, we study the problem of attentional processing for graph classification. The use of attention allows us to focus on small but informative parts of the graph, avoiding noise in the rest of the graph. We present a novel RNN model, called the \textit{Graph Attention Model} (\method), that processes only a portion of the graph by adaptively selecting a sequence of ``interesting" nodes. The model is equipped with an external memory component which allows it to integrate information gathered from different parts of the graph. We demonstrate the effectiveness of the model through various experiments.
\end{abstract}

\section{Introduction}
\label{sec:intro}
Graph-structured data arise naturally in a wide variety of applications including bioinformatics \cite{Borgwardt05}, chemoinformatics \cite{Duvenaud15}, social network analysis \cite{Backstrom11}, urban computing \cite{Bao17}, and cyber-security \cite{Chau11}. In many cases, the primary task is identifying the class labels of the graphs in a dataset. In chemoinformatics, for instance, molecules can be represented as graphs, where nodes correspond to atoms and each edge signifies the presence of a chemical bond between a pair of atoms. The task then is to predict the label of each graph -- for instance, the anti-cancer activity or toxicity of a molecule. To solve this problem, the usual strategy is to calculate certain graph statistics that will help in discriminating between the different types of graphs. We do this because we can expect graphs belonging to a particular class to exhibit some common behavior that is not typically observed among the other graphs. 

\begin{figure*}[t]
\centering
\includegraphics[width=0.8\linewidth]{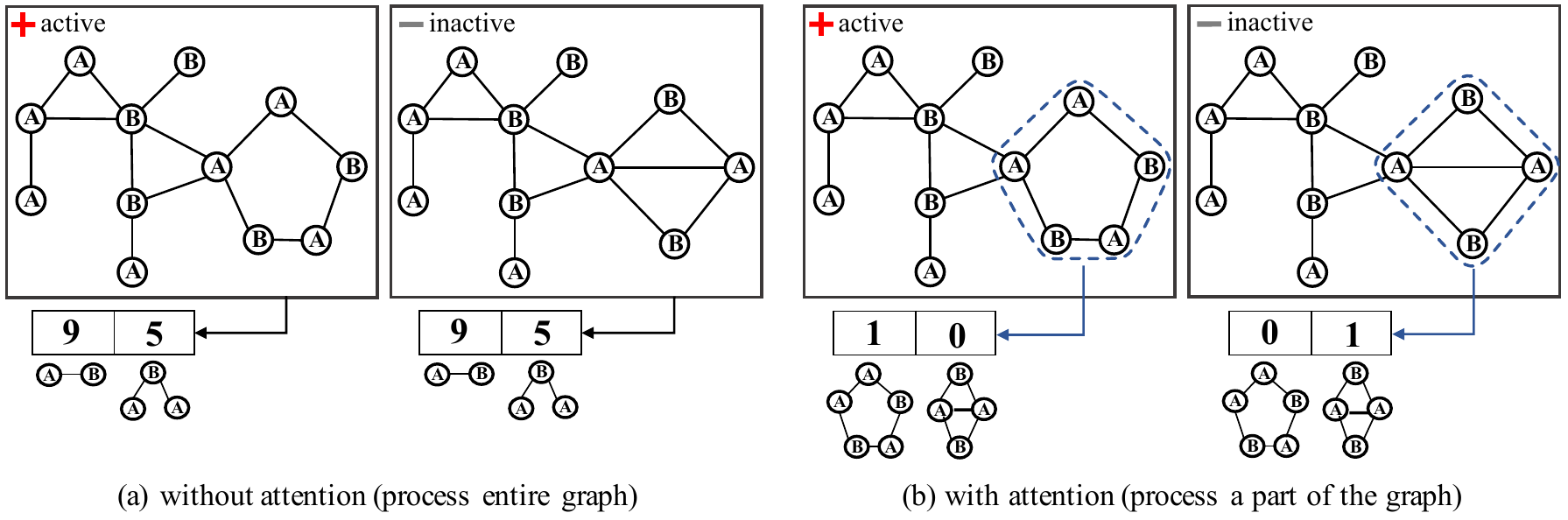}
\caption{(a) When processing an entire graph to get the count of various subgraph patterns, we are often limited to counting relatively simple patterns since the total number of patterns can grow exponentially with the size of the patterns \cite{Rossi17}. Given a large and noisy graph with relatively complex patterns, this approach fails. (b) Attention can be used to allow us to focus on informative parts of the graph only, helping to uncover the more complex and useful patterns. This allows us to compute graph representations that can better discriminate between positive and negative samples.}
\label{fig:diff}
\end{figure*}

Currently, in most existing solutions, the entire graph is taken into account when calculating such statistics. The Morgan algorithm for calculating circular fingerprints (\textit{i.e.} graph representation), follows an iterative process which recomputes each node's attribute vector by concatenating and hashing the attributes of neighboring nodes \cite{Rogers10}. The final graph representation is then computed using the new attributes of all the nodes in the graph. Another popular technique is the random walk graph kernel, which computes the number of common walks in a pair of graphs to measure graph similarity \cite{Vishy10}. This can be done on the product graph of two graphs and again the entire graphs are considered. Because the entire graph has to be processed, it is usually costly if not infeasible, to compute representations of large real-world graphs \cite{Rossi17}. 


On top of the significant computational cost that is incurred, processing the entire graph can also have a negative impact on the overall performance of a model on graph classification. This is particularly true if the significant subgraph patterns for a given task are sparse and confined to small neighborhoods within the graph. Since the rest of the graph do not contain anything that will help identify graph label, processing the entire graph can inadvertently cause noise to be introduced. An example of this is illustrated in Figure~\ref{fig:diff}.

To address the issues mentioned above, we study a model that uses attention to actively select a region in the graph to process. By using attention to focus on informative parts of the graph, we are able to improve the model's performance while keeping the computation cost (space, in particular) low. This is particularly true on graphs where the signal-to-noise ratio is significant since attention allows us to ignore noisy parts of the graph. For instance, when studying the interaction networks of complex diseases, researchers have found that it is often beneficial to focus on specific subnetworks that are associated with the disease \cite{Cho12}. Inspired by the recent success of Recurrent Neural Networks (RNN) with attention on vision-related tasks \cite{Minh14}, we explore an RNN model with a built-in attention mechanism for graph-structured data. 


\section{Related Work} 
\label{sec:related-work}
Many different techniques have been proposed to solve the graph classification problem. One popular approach is to use a graph kernel to measure similarity between different graphs \cite{Niko17}. This similarity can be measured by considering various structural properties like the shortest paths between nodes \cite{Borgwardt052}, the occurrence of certain graphlets or subgraphs \cite{Shervashidze09}, and even the structure of the graph at different scales \cite{Kondor16}. Recently, several new methods, which generalize over previous approaches, have been introduced. These methods use a deep learning framework to learn data-driven representations \cite{Yanardag15,Duvenaud15}. One thing that is common among all these approaches is that the entire graph is processed to compute the final representation. In contrast, the model we study only processes a portion of the graph and attention is used to determine parts of the graph to focus on.

Deep learning frameworks equipped with attentional processing have been shown to perform well in a variety of tasks. In \cite{Luong15}, attention was used to allow the model to attend to a subset of the source words in the language translation task. Meanwhile, \cite{Xu15} used attention to help a model fix its gaze on salient objects for image captioning and \cite{Minh14} applied attention to the image classification task. \cite{Chen15}, on the other hand, used attention to guide a CNN to focus on relevant objects for the visual question answering task. Although attentional processing has been applied successfully to many problems, most of the existing work lie in the computer vision or natural language processing domains. Recently, a model was introduced that explores attentional processing on medical ontology graphs \cite{Choi17}. However, our work is significantly different from the latter as the model in \cite{Choi17} is specifically designed for medical ontologies and work on directed acyclic graphs (DAG) while we explore an attention mechanism on general attributed graphs. To the best of our knowledge, this is the first work that explores attention on general graph-structured data.

Finally, we also experiment with an architecture that has a simple external memory to allow multiple agents to integrate information from various parts of the graph. In a sense, this is conceptually similar to the memory networks of \cite{Sukhbaatar15,Prakash17}.

\begin{figure*}[t]
\centering
\includegraphics[width=0.94\linewidth]{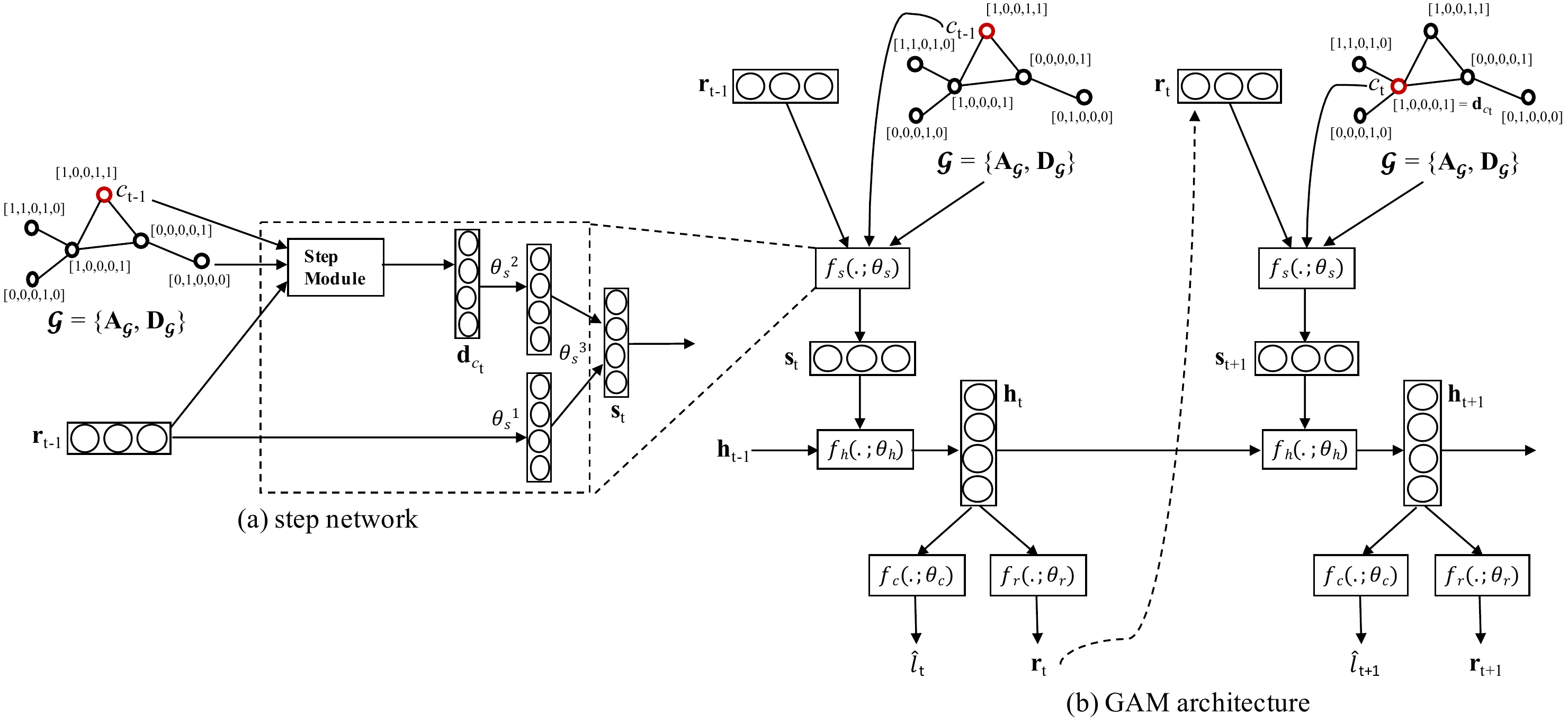}
\caption{{\bf(a) Step network: }Given a labeled graph $\mathcal{G}$ (composed of the adjacency matrix $\mathbf{A}_\mathcal{G}$, and the attribute matrix $\mathbf{D}_\mathcal{G}$), a current node $c_{t-1}$, and a stochastic rank vector $\mathbf{r}_{t-1}$, the step module takes a step from the current node $c_{t-1}$ to one of its neighbors $c_t$, prioritizing those whose type (\textit{i.e.}, node label) have higher rank in $\mathbf{r}_{t-1}$. The attribute vector of $c_t$, ${\bf d}_{c_t}$, is extracted and mapped to a hidden space using the linear layer parameterized by $\theta_s^2$. Similarly, ${\bf r}_{t-1}$ is mapped using another linear layer parameterized by $\theta_s^1$. Information from these two sources are then combined using a linear layer parameterized by $\theta_s^3$ to produce ${\bf s}_t$, or the step embedding vector which represents information captured from the current step we took. {\bf(b) \method { architecture:} }We use an RNN as the core component of the model; in particular, we use the Long Short-Term Memory (LSTM) variant \cite{Gers99}. At each time step, the core network $f_h(.;\theta_h)$ takes the step embedding $\mathbf{s}_t$ and the internal representation of the model's history from the previous step $\mathbf{h}_{t-1}$ as input, and produces the new history vector $\mathbf{h}_t$. The history vector $\mathbf{h}_t$ can be thought of as a representation or summary of the information we've aggregated from our exploration of the graph thus far. The rank network $f_r(.;\theta_r)$ uses this to decide which types of nodes are more ``interesting" and should thus be prioritized in future exploration. Likewise, the classification network $f_c(.;\theta_c)$ uses $\mathbf{h}_t$ to make a prediction on the graph label.}
\label{fig:banner1}
\end{figure*}

\section{Graph Attention Model}
\label{sec:framework}
To simplify the discussion, we begin by describing a basic attention model. In subsequent discussion, we introduce a variant with more refined attention and external memory. 

Although the proposed framework is general and can be adopted for a variety of tasks, we choose to frame the discussion in the context of graph classification on attributed graphs. More formally, given a set of attributed graphs $\mathcal{D} = \{(\mathcal{G}_1, \ell_1), (\mathcal{G}_2, \ell_2), \cdots, (\mathcal{G}_n, \ell_n)\}$, the goal is to learn a function $f: \mathbb{G} \rightarrow \mathcal{L}$, where $\mathbb{G}$ is the input space of graphs and $\mathcal{L}$ is the set of graph labels. Here each graph $\mathcal{G}_i = (\mathbf{A}_{\mathcal{G}_i}, \mathbf{D}_{\mathcal{G}_i})$ is comprised of an adjacency matrix $\mathbf{A}_{\mathcal{G}_i} \in \mathbb{N}^{N_i \times N_i}$, and an attribute matrix $\mathbf{D}_{\mathcal{G}_i} \in \mathbb{R}^{N_i \times D}$, where $N_i$ is the number of nodes in graph $i$ and $D$ is the number of attributes. Each graph also has a corresponding label $\ell_{\mathcal{G}_i}$.

In this work, we formulate the problem of applying attention on graph-structured data as a decision process of a goal-directed agent traversing along an input attributed graph. The agent starts at a random node on the graph and, at each time step, moves to a neighboring node. The information available to the agent is limited to the node it chooses to explore. Since global information about the graph is unavailable, the agent needs to integrate information over time to help it determine the parts of the graph to explore further. The ultimate goal of the agent is to collect enough information that will allow it to make a correct prediction on the label of the graph. 

The agent will only explore a small portion of the graph with the attention mechanism guiding it in its exploration. If the graph is large, we can also initialize multiple agents at different nodes in the graph and run them in parallel. Deploying multiple agents can help improve the performance of the model since each agent can explore a different part of the graph with attention helping to steer each agent's exploration along the local neighborhood. This allows us to use the model on large graphs that may be difficult or impossible to load into memory. 

\subsection{Proposed Model}
Our proposed model has an RNN at its core, as shown in Figure~\ref{fig:banner1}. At each time step, the core network processes new information from the step that was just taken and integrates this into its internal representation together with information retained from previous steps. It uses this information to predict the label of the input graph and to decide which areas of the graph to prioritize for further exploration in the next time step.

{\bf Step module: }At each time step, the step module considers the one-hop neighborhood of the current node $c_{t-1}$ and picks a neighbor $c_t$ to take a step towards. The step module is biased towards picking neighbors whose types or labels have higher rankings in the rank vector $\mathbf{r}_{t-1}$. The attribute vector of the chosen node is then extracted and fed together with $\mathbf{r}_{t-1}$ to produce the step representation $\mathbf{s}_t = f_s(\mathbf{d}_{c_t}, \mathbf{r}_{t-1};\theta_s)$ (see Figure~\ref{fig:banner1}a). The step representation $\mathbf{s}_t$ is the new information available to the core LSTM network at each time step. The step algorithm is summarized in Algorithm~\ref{alg:step}.

{\bf Node type: }The way we label or assign types to nodes allows us to bias the exploration towards certain nodes at different stages of the exploration. Depending on the application, the node type can be a simple discrete value (\textit{e.g.}, type of atom in a molecular graph) or it can be something more elaborate like a category derived from log-binning several attributes that capture the local structure of the node. We give a simple example of the latter case. Suppose the agent wants to visit one of two Carbon nodes adjacent to it, it cannot differentiate between the nodes under the first node typing strategy. In the second method, the node type may be calculated based on the statistics encoded in the $k$-hop neighborhood of each node and this allows us to differentiate between the two Carbon nodes. Using more complex node typing strategies may, however, increase the number of node types substantially and one may have to look into reinforcement learning strategies that work well when the discrete action space is large \cite{Dulac15}.

{\bf History: }The core LSTM network maintains a history vector which is a summary of all the information obtained by the agent in its exploration of the graph thus far. At each time step, as new information becomes available in the form of $\mathbf{s}_t$ from the step we just took, the history vector is updated via $\mathbf{h}_t = f_h(\mathbf{s}_t, \mathbf{h}_{t-1};\theta_h)$. This allows the core network to integrate information over time. 

We use an LSTM in our architecture as it is superior to simple RNNs in capturing long-range dependencies. Even though LSTMs have a more sophisticated memory model when compared to simple RNNs, it has been shown that they still have trouble remembering information that was inputted too far in the past \cite{Weston14}. Because of this, on large graphs, it may be better to deploy multiple agents with each agent exploring a relatively small neighborhood rather than having one agent traverse the graph for a long period. To integrate information, we can augment the architecture with a shared external memory \cite{Sukhbaatar15}. Additionally, a network conditioned on the current history vector can be trained to allow the model to selectively save information to memory. This will allow the model to store information that is useful for graph classification ({\textit{e.g.}, discriminative subgraphs).

\begin{algorithm}[t]
\begin{algorithmic}[1]
\Procedure{Step}{$\mathbf{r}_{t-1} \in \mathbb{R}^{R}, \mathbf{A} \in \mathbb{N}^{N \times N}, \mathbf{D} \in \mathbb{R}^{N \times D}, c_{t-1}$}

\State $\mathbf{a} \gets  \mathbf{A}[c_{t-1}, :\,]$
\State $\mathbf{T} \gets \tau(\mathbf{D})$ \hspace{1pt} \Comment{$\mathbf{T} \in \mathbb{R}^{N \times R}$ is a matrix of one-hot row vectors indicating node types; we assume that type can be derived from node attributes.}
\State $\mathbf{p} \gets (\mathbf{T}\,\mathbf{r}_{t-1})^\top$
\State $\mathbf{p} \gets \mathbf{p} \odot \mathbf{a}$
\State $\mathit{d} \gets \sum_{i} p_i$
\State $\mathbf{p} \gets \mathbf{p} \odot \frac{1}{\mathit{d}}$

\State $c_{t} \sim $ Multinomial$(\mathbf{\pi} = \mathbf{p})$ \hspace{1pt} \Comment{Sample a neighbor from multinomial distribution parameterized by $\mathbf{p}$.}




\State \textbf{return} $\mathbf{D}[c_t, :\,], c_t$
\EndProcedure
\end{algorithmic}
\caption{Procedure to pick a neighbor to move to. The algorithm is biased towards picking neighbors whose types have higher ranks in $\mathbf{r}_{t-1}$. Here, $\odot$ represents element-wise multiplication and $\triangleright$ denotes the start of a comment.}
\label{alg:step}
\end{algorithm}

{\bf Actions: }Given the new history vector that captures what the agent has seen so far, the agent performs two actions at each time step. First, it predicts the label of the input graph $\hat{l}_t = \operatornamewithlimits{arg\,max}\limits_i P(y = i | f_c(\mathbf{h}_t;\theta_c))$ from the softmax output of the classification network conditioned on $\mathbf{h}_t$. Second, it uses the rank network to generate the rank vector $\mathbf{r}_t = f_r(\mathbf{h}_t;\theta_r)$ that will help ``steer" exploration in the next step by ranking the importance of different types of nodes.

Primarily, the rank vector's job is to encode the importance of different types of nodes. However, we can augment it to include additional actions such as one for deciding when to stop further exploration if the agent is confident it has enough information to classify the graph correctly. Another possible action is the one that allows the model to transfer its current internal information to a memory component.

{\bf Reward: }In the typical reinforcement learning setting, the agent receives new information $x_{t+1}$ from the environment and a reward signal $r_{t+1}$ after taking an action at each time step $t$. The goal of the agent is to maximize the reward it receives which is usually quite sparse and delayed: $R = \sum_{t=1}^T r_t$. In our setting, $\mathbf{x}_{t+1} = \mathbf{d}_{c_{t+1}}$ and the reward is given only at the end, where $r_T = 1$ if the model classified the graph correctly and $r_T = -1$ otherwise. Hence $R = r_T$.

Under this formulation, we have what can be considered a Partially Observable Markov Decision Process (POMDP). In this setting, we only obtain partial information about the graph or our environment through our interactions with it at each time step. As in \cite{Minh14}, our goal is to learn a policy $\pi((\mathbf{r}_t, \hat{l}_t|s_{1:t};\theta))$ with parameters $\theta$ that maps the sequence of our past interactions with the environment $s_{1:t} = \mathbf{x}_1, \mathbf{r}_1, \hat{l}_1, \cdots, \mathbf{x}_{t-1}, \mathbf{r}_{t-1}, \hat{l}_{t-1}, \mathbf{x}_t$ to a distribution over actions for the current time step $t$. In other words, given the history of past interactions as summarized in the history vector $\mathbf{h}_t$, the classification network $f_c(.;\theta_c)$ and the rank network $f_r(.;\theta_r)$ -- or our policy networks -- learn to generate actions that maximize reward.

\subsection{Training}
\label{sec:training}
Together, the core LSTM network, the step network, and the rank network work in conjunction with each other to form the policy of the agent. We learn the parameters $\theta = \{\theta_h, \theta_s, \theta_r \}$ of these networks to maximize the total reward the agent can expect to obtain. Since each specific policy for the agent induces a distribution over the possible interaction sequences $s_{1:T}$, we want to train our policy to maximize the reward under the generated distribution: $J(\theta) = \mathbb{E}_{P(s_{1:T};\theta)}[R]$. 

It is a non-trivial task to maximize $J$ exactly as we are dealing with a very large, and possibly infinite, number of possible interaction sequences. However, since we frame the problem as a POMDP, we are able to obtain a sample approximation of the gradient of $J$ by using the technique introduced by \cite{Williams92} as shown in \cite{Minh14}. This is given by
\begin{align}
\label{eq:reinforce}
\nabla_\theta J \approx \frac{1}{M} \sum\limits_{i=1}^M \sum\limits_{t=1}^{T-1} \nabla_\theta \text{log}\,\pi(\mathbf{r}_t^i[\tau(c_{t+1}^i)] | s_{1:t}^i;\theta) \gamma^{T-t} R^i
\end{align}
where the $s^i$'s are the interaction sequences from running the agent under the current policy for $i = 1, \cdots,M$ episodes, $\gamma \in (0,1]$ is a discount factor that allows us to attribute more significance to actions performed closer to time $T$ or when the prediction was made, and $\tau(c_{t+1}^i)$ is a function that maps a node to its type. The intuition behind equation~\ref{eq:reinforce}, which is also known as the REINFORCE rule, is as follows. We run the agent with the current policy to obtain samples of interaction sequences. The parameters $\theta$ are then adjusted to increase the log-probability or rank of the type of nodes that were frequently selected during episodes that resulted in a correct prediction. Training the policy this way allows us to increase the chance that the agent will choose to take a step towards a particular type of node the next time it finds itself in a similar state. To compute $\nabla_\theta \text{log}\,\pi(\mathbf{r}_t^i[\tau(c_{t+1}^i)] | s_{1:t}^i;\theta)$, we simply compute the gradient of our network at each time step, this can be done using standard backpropagation \cite{Wierstra07}. Note that we only adjust the log-probabilities for $t = 1, \cdots,T-1$ since the rank vector $\mathbf{r}_t$ in the last step is no longer used.

Since the gradient estimate in Equation~\ref{eq:reinforce} may exhibit high variance, one may choose to estimate $\nabla_\theta J$ via 
\begin{align}
\label{eq:reinforce2}
\frac{1}{M} \sum\limits_{i=1}^M \sum\limits_{t=1}^{T-1} \nabla_\theta \text{log}\,\pi(\mathbf{r}_t^i[\tau(c_{t+1}^i)] | s_{1:t}^i;\theta) (\gamma^{T-t} R^i - b_t^i)
\end{align}
instead. This provides us with an estimate that is equal in expectation to the original formulation but with possibly lower variance \cite{Minh14}. Here $b_t^i = f_b(s_{1:t}^i;\theta_b) = f_b(h_t^i;\theta_b)$ captures the cumulative reward we can expect to receive for a state $h_t^i$. The term $(\gamma^{T-t} R^i - b_t^i)$, or the advantage of choosing an action, allows us to increase the log-probability of actions that resulted in a much larger expected cumulative reward and to decrease the log-probability of actions that resulted in the reverse. We can train the parameter $\theta_b$ of $f_b$ by reducing the mean squared error of $R^i - b_t^i$.

Finally, we use cross entropy loss to train the classification network $f_c(.;\theta_c)$ by maximizing $\text{log}\,\pi(l_T | s_{1:T}; \theta_c)$, where $l_T$ is the true label of the input graph $\mathcal{G}$. As in \cite{Minh14}, we use this hybrid loss formulation where the rank network $f_r$ is trained at each time step using REINFORCE and the classification network $f_c$ and the baseline network $f_b$ are trained using the classical approach from supervised learning.

\subsection{Space Complexity}
Let $\triangle_{\mathcal{G}}$ be the max node degree for graph $\mathcal{G}$ and $D$ be the dimension of the node attribute vector. Since the agent only moves to one of the current node's neighbors at each time step, we only need to store a $\triangle_{\mathcal{G}} \times D$ matrix containing the attributes of neighboring nodes at any given time. After taking a step to a new node, the attribute matrix for the new set of neighbors can be fetched from disk. Ignoring the space needed to store $\mathbf{r}$, $\mathbf{s}$, $\mathbf{h}$, $c$, and the parameters of our model, which are constant and negligible, our model has a space complexity of $\mathcal{O}(\triangle_{\mathcal{G}}D)$ which is quite small in practice.

\subsection{Initialization}
\label{sec:init}
For each new instance, we initialize the start vertex $c_0$ by selecting a random node in the input graph and the rank vector $\mathbf{r}_0$ is initialized to the uniform distribution.

\subsection{Attention with Memory} 
\label{sec:mem}
When predicting the label of an input graph, one may choose to average the softmax output of several runs by initializing multiple agents at different starting locations in the graph. In this case, we can view each agent as one classifier in an ensemble where we predict by voting. While averaging the predictions of several agents can certainly improve classification performance, our model is still at a disadvantage against methods that integrate information from the entire graph. This is because each agent makes a prediction independently, using only the information it gathered from a local area within the graph.

To remedy this, we introduce a variant of our model with a shared external memory component that can store information from multiple agents. In this architecture, each agent $i$ for $i = 1, \cdots, n$ stores information in a local memory component $\mathbf{p}_i$, these are then combined to form the shared memory $\mathbf{m}$ that the classification network uses to make a single prediction. In the simplest case, $\mathbf{p}_i = \mathbf{h}_T^{i}$, which means we use the final history vector as each agent's local memory. However, not all parts of an agent's walk through the graph may yield equally important information. To allow the model to retain only information useful to the task we set $\mathbf{p}_i = \sum_{j = 1}^{T} u^i_j \mathbf{h}_j^{i}$, where the $u^i_j$'s are the softmaxed output of $f_u(h_j^i;\theta_u)$ which decides how useful a particular ``piece of memory" is. In other words, we do weighted pooling to obtain our local memory. This can be viewed as another form of attention. Finally, to integrate information from multiple agents, we simply set $\mathbf{m} = \frac{1}{n} \sum_{i=1}^n \mathbf{p}_i$. This modification allows us to integrate information from various regions in the graph and is especially helpful if the graph is large and we only take a small number of steps $T$. Note that each agent's exploration is still guided by the attention mechanism proposed earlier.

Various modifications can be made to this architecture. For instance, we can choose to condition the output of the rank network on the local memory or even the shared external memory. Additional actions can also be introduced to allow the model to modify or rewrite the shared memory. In this work, however, we choose to test on the simplest version to demonstrate its efficacy. Due to space limitations, we do not show a diagram of the model with memory and instead include it as supplemental material.

\begin{figure}[t]
\centering
\includegraphics[width=0.92\linewidth]{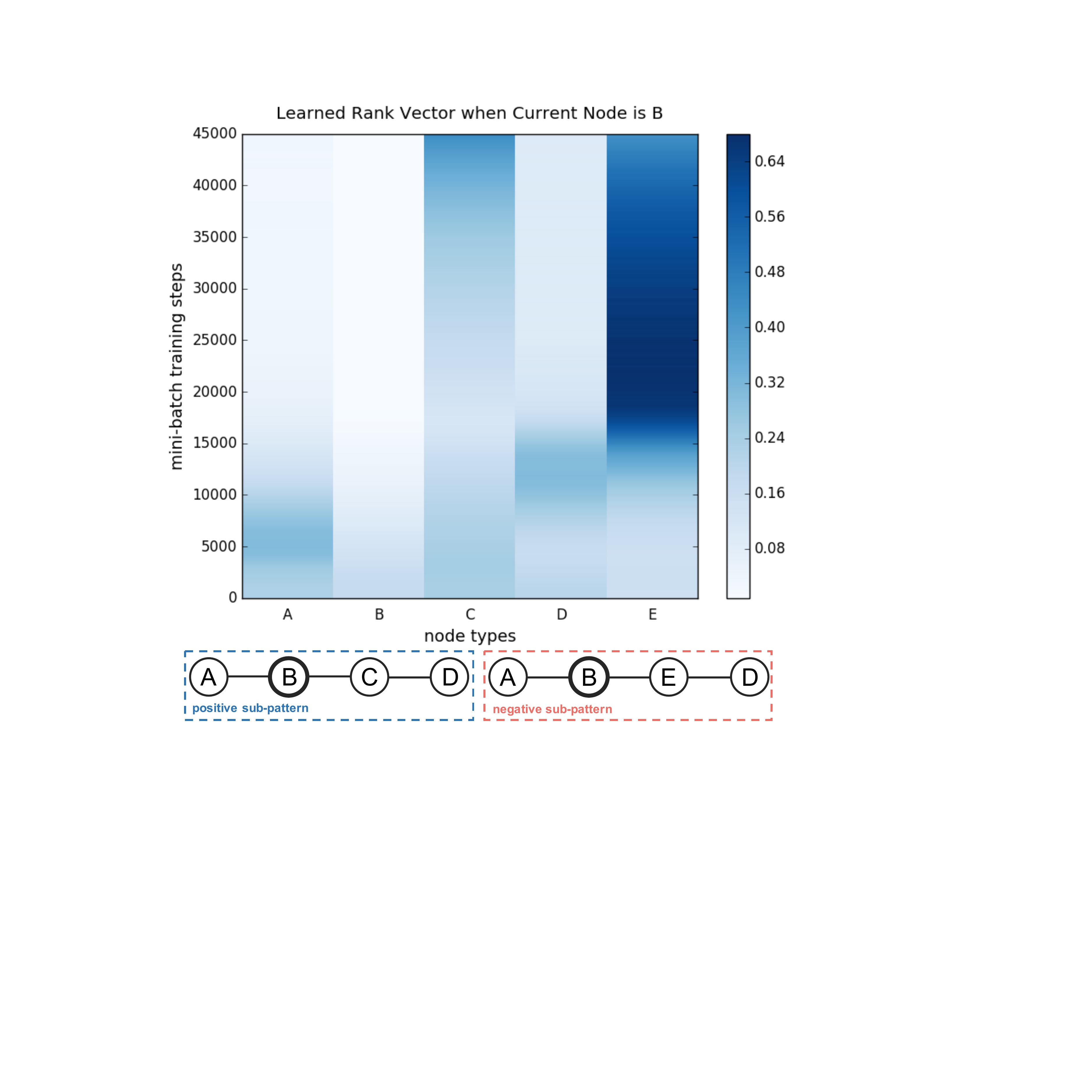}
\caption{Rank values, over time, in the generated rank vector $\mathbf{r}_1$ when the rank network is given $\mathbf{h}_1$ encoding information from an initial step onto node $B$. Higher rank value signifies more importance.}
\label{fig:motiv}
\end{figure}

\begin{table*}[t]
\caption{Summary of experimental results: ``average accuracy $\pm$ SD ({\color{blue} rank})". The ``ave. rank" column shows the average rank of each method. The lower the average rank, the better the overall performance of the method.}
\label{tab:acc}
\begin{center}
\begin{tabular}{l l l l l l l l l l}
\hline
\multicolumn{1}{c}{\multirow{2}{*}{\bf method}}  & \multicolumn{5}{c}{\rule{0pt}{2ex} \bf dataset} & \multicolumn{1}{c}{\multirow{2}{*}{\bf \shortstack{ave.\\rank}}}\\
\cline{2-6}
& \multicolumn{1}{c}{\rule{0pt}{2ex} HIV} & \multicolumn{1}{c}{NCI-1} & \multicolumn{1}{c}{NCI-33} & \multicolumn{1}{c}{NCI-83} & \multicolumn{1}{c}{NCI-123} & \\
\hline
{\rule{0pt}{2ex}Agg-Attr}& 69.58 $\pm$ 0.03 ({\color{blue} 4}) & 64.79 $\pm$ 0.04 ({\color{blue} 4}) & 61.25 $\pm$ 0.03 ({\color{blue} 6}) & 58.75 $\pm$ 0.05 ({\color{blue} 6}) & 60.00 $\pm$ 0.02 ({\color{blue} 6}) & \multicolumn{1}{c}{\color{blue} 5.2}\\
Agg-WL& 69.37 $\pm$ 0.03 ({\color{blue} 6}) & 62.71 $\pm$ 0.04 ({\color{blue} 6}) & 67.08 $\pm$ 0.04 ({\color{blue} 5}) & 60.62 $\pm$ 0.02 ({\color{blue} 4}) & 62.08 $\pm$ 0.03 ({\color{blue} 5}) & \multicolumn{1}{c}{\color{blue} 5.2}\\
Kernel-SP& 69.58 $\pm$ 0.04 ({\color{blue} 4}) & 65.83 $\pm$ 0.05 ({\color{blue} 3}) & 71.46 $\pm$ 0.03 ({\color{blue} 1}) & 60.42 $\pm$ 0.04 ({\color{blue} 5}) & 62.92 $\pm$ 0.07 ({\color{blue} 4}) & \multicolumn{1}{c}{\color{blue} 3.4}\\
Kernel-Gr& 71.88 $\pm$ 0.05 ({\color{blue} 3}) & 67.71 $\pm$ 0.06 ({\color{blue} 1}) & 69.17 $\pm$ 0.03 ({\color{blue} 3}) & 66.04 $\pm$ 0.03 ({\color{blue} 3}) & 65.21 $\pm$ 0.05 ({\color{blue} 2}) & \multicolumn{1}{c}{\color{blue} 2.4}\\
\method& 74.79 $\pm$ 0.02 ({\color{blue} 2}) & 64.17 $\pm$ 0.05 ({\color{blue} 5}) & 67.29 $\pm$ 0.02 ({\color{blue} 4}) & 67.71 $\pm$ 0.03 ({\color{blue} 2}) & 64.79 $\pm$ 0.02 ({\color{blue} 3}) & \multicolumn{1}{c}{\color{blue} 3.2}\\
\methodmem& 78.54 $\pm$ 0.04 ({\color{blue} 1}) & 67.71 $\pm$ 0.04 ({\color{blue} 1}) & 69.58 $\pm$ 0.02 ({\color{blue} 2}) & 70.42 $\pm$ 0.03 ({\color{blue} 1}) & 67.08 $\pm$ 0.03 ({\color{blue} 1}) & \multicolumn{1}{c}{\color{blue} 1.2}\\
\hline
\end{tabular}
\end{center}
\end{table*}

\section{Experiments} \label{sec:exp}

\subsection{Motivating Example}
Before we consider the details of our main experimental setup, we introduce a simple motivating example that shows how attention can be used to guide an agent towards more relevant regions in the graph. For this toy example, we generated a small dataset of random graphs. We embedded several patterns or subgraphs in the generated graphs, two of which were the $3$-paths $A-B-C-D$, and $A-B-E-D$. The former pattern was embedded primarily onto positive samples while the latter was included in negative samples. In Figure~\ref{fig:motiv}, we show the output of the rank network, over time, when it is given the history vector $\mathbf{h_1}$ capturing the initial step onto the node of type $B$. It is interesting to note that, initially, the rank network assigns more or less equal importance to the five types of nodes. However, after some time, it learns to prioritize the nodes of types $C$, and $E$. This guarantees that the agent will prioritize exploration in the right direction, giving the model enough information to classify the graphs correctly in a small number of steps.

\subsection{Experimental Setup} \label{sec:exp-settings}
\subsubsection{Data} \label{sec:exp-data}
We evaluated our proposed method on the binary classification task using five molecular graph datasets: HIV, NCI-1, NCI-33, NCI-83, and NCI-123. Since the molecular structures in the datasets were encoded using the SMILES format \cite{Weininger88}, we used the RDKit\footnote{http://www.rdkit.org/} package to convert each string into its corresponding graph. We used the same package to extract the following information for each node ({\textit i.e.} atom) to use as node attributes: atom element, node degree, total number of attached hydrogens, the implicit valence, and atom aromaticity. Atom element was used to label or assign types to the nodes. The graph class labels indicate the anti-cancer property (active or negative) of each molecule. The datasets are all highly imbalanced with far more negative samples than positive ones. Like previous work \cite{Kong11,Yanardag15}, we balanced the datasets. All experiments are conducted on balanced subsets containing 500 randomly selected graphs. 

\subsubsection{Compared Methods}
In order to demonstrate the effectiveness of our proposed approach, we compare it against several baseline methods, all of which utilize the entire graph for feature extraction. To the best of our knowledge, this is the first work on attention with graphs so we compare against baselines that observe the entire graph which puts our model (\method) at a disadvantage since it only has partial observability. The compared methods are summarized below.
\begin{itemize}
\setlength{\itemsep}{1pt}
\setlength{\parskip}{0pt}
\setlength{\parsep}{0pt}
	\item \textbf{Agg-Attr}: Given an attributed graph, one simple way to construct a feature vector is to get the component-wise average of the attribute vectors of all the nodes in the graph.
	\item \textbf{Agg-WL}: The first approach captures information from node attributes. However, it completely ignores the graph's structural information. The second method uses the Weisfeiler-Lehman (WL) algorithm \cite{Shervashidze11} to calculate new node attributes that capture the local neighborhood of each node. The algorithm works by iteratively assigning a new attribute to each node by computing a hash of the attributes of neighboring nodes. We simply average the new attributes after running the WL algorithm to use as feature vector used for prediction.
	\item \textbf{Kernel-SP}: As in \cite{Yanardag15}, we compare against the shortest path (SP) kernel which measures the similarity of a pair of graphs by comparing the distance of the shortest paths between nodes in the graphs. Since we use attributed graphs, we label the nodes in the graph by concatenating the categorical attributes.
	\item \textbf{Kernel-Gr}: As in \cite{Yanardag15}, we also compare against the graphlet kernel which measures graph similarity by counting the number of different graphlets. Here, we evaluate against the 3-graphlet kernel and nodes are labeled in the same way as above.
	\item \textbf{\method}: Our proposed approach which uses attention to steer the walk of an agent on an input graph.
	\item \textbf{\methodmem}: Proposed approach with external memory.
\end{itemize}

\begin{table*}[t]
\caption{Performance of the baselines when we restrict their setting to that of {\method } where they are given 20 randomly selected partial snapshots of each graph and have to predict by voting. The column ``full" indicates the performance when the entire graph is seen and ``partial" shows the performance when only parts of the graph is seen. ``Diff." is the difference in performance, a ${\color{red} \downarrow}$ means that performance deteriorated when only partial information is available and ${\color{blue} \uparrow}$ shows increase in performance.}
\label{tab:active}
\begin{center}
\begin{tabular}{|l |l l l |l l l |l l l |l l l |l l l|}
\hline
\multicolumn{1}{|c|}{\multirow{3}{*}{\bf \scriptsize method}}  & \multicolumn{15}{c|}{\bf \scriptsize dataset}\\
\cline{2-16}
{\rule{0pt}{1ex}} & \multicolumn{3}{c|}{\scriptsize HIV} & \multicolumn{3}{c|}{\scriptsize NCI-1} & \multicolumn{3}{c|}{\scriptsize NCI-33} & \multicolumn{3}{c|}{\scriptsize NCI-83} & \multicolumn{3}{c|}{\scriptsize NCI-123} \\
\cline{2-16}
{\rule{0pt}{1ex}} & \multicolumn{1}{c}{\scriptsize full} & {\scriptsize partial} & \multicolumn{1}{c|}{\scriptsize diff.} & \multicolumn{1}{c}{\scriptsize full} & {\scriptsize partial} & \multicolumn{1}{c|}{\scriptsize diff.} & \multicolumn{1}{c}{\scriptsize full} & {\scriptsize partial} & \multicolumn{1}{c|}{\scriptsize diff.} & \multicolumn{1}{c}{\scriptsize full} & {\scriptsize partial} & \multicolumn{1}{c|}{\scriptsize diff.} & \multicolumn{1}{c}{\scriptsize full} & {\scriptsize partial} & \multicolumn{1}{c|}{\scriptsize diff.}\\
\hline
{\rule{0pt}{2ex}\scriptsize Agg-Attr} & {\scriptsize 69.58} & {\scriptsize 64.17} & {\scriptsize 05.41 {\tiny(${\color{red} \downarrow}$)}} & {\scriptsize 64.79} & {\scriptsize 59.58} & {\scriptsize 05.21 {\tiny(${\color{red} \downarrow}$)}} & {\scriptsize 61.25} & {\scriptsize 58.54} & {\scriptsize 02.71 {\tiny(${\color{red} \downarrow}$)}} & {\scriptsize 58.75} & {\scriptsize 62.71} & {\scriptsize 03.96 {\tiny(${\color{blue} \uparrow}$)}} & {\scriptsize 60.00} & {\scriptsize 57.50} & {\scriptsize 02.50 {\tiny(${\color{red} \downarrow}$)}} \\
{\scriptsize Agg-WL} & {\scriptsize 69.37} & {\scriptsize 56.04} & {\scriptsize 13.33 {\tiny(${\color{red} \downarrow}$)}} & {\scriptsize 62.71} & {\scriptsize 51.46} & {\scriptsize 11.25 {\tiny(${\color{red} \downarrow}$)}} & {\scriptsize 67.08} & {\scriptsize 49.79} & {\scriptsize 17.29 {\tiny(${\color{red} \downarrow}$)}} & {\scriptsize 60.62} & {\scriptsize 51.46} & {\scriptsize 09.16 {\tiny(${\color{red} \downarrow}$)}} & {\scriptsize 62.08} & {\scriptsize 52.29} & {\scriptsize 09.79 {\tiny(${\color{red} \downarrow}$)}}\\
{\scriptsize \method} & \multicolumn{1}{c}{\scriptsize -} & {\scriptsize 74.79} & \multicolumn{1}{c|}{\scriptsize -} & \multicolumn{1}{c}{\scriptsize -} & {\scriptsize 64.17} & \multicolumn{1}{c|}{\scriptsize -} & \multicolumn{1}{c}{\scriptsize -} & {\scriptsize 67.29} & \multicolumn{1}{c|}{\scriptsize -} & \multicolumn{1}{c}{\scriptsize -} & {\scriptsize 67.71} & \multicolumn{1}{c|}{\scriptsize -} & \multicolumn{1}{c}{\scriptsize -} & {\scriptsize 64.79} & \multicolumn{1}{c|}{\scriptsize -}\\
\hline
\end{tabular}
\end{center}
\end{table*}

We used a logistic regression (LR) classifier with the first two baselines. To reduce overfitting, we applied $\ell_1$ and $\ell_2$ regularization and used a grid search over $\{0.01, 0.1, 1.0\}$ to select the ideal regularization penalty. Furthermore, we also did a grid search over the number of iterations for the WL algorithm, we tested over $\{2, 3, 4\}$. For a fair comparison, we limited the classification network for both our methods to a single softmax layer to make it equivalent to LR. We also limited the number of hidden layers in all other networks of our model to a single layer, whenever possible. For the graph-kernel based approaches, we used an SVM classifier using the precomputed kernel generated by each approach. Here, we did a grid search over $C = \{0.01, 0.1, 1.0\}$. We used a vector in $\mathbb{R}^{200}$ for Agg-WL and limited the size of the LSTM history vector to this size as well. In particular, we tried size = $\{156, 200\}$. We also tried the following sizes for the first and second hidden layers, respectively, of the step network: $(128, 164)$, and $(64, 128)$.

Since we did not find any noticeable change in the performance of {\method } when increasing the following parameters, we fixed their values. We set the number of steps $T = 12$ and the number of samples $M = 20$. $M$ is also the number of agents we run on each graph for prediction. For \methodmem, we did a grid search over $T = \{12, 25\}$, and $M = \{5, 10\}$. We use the Adam algorithm for optimization \cite{Kingma15} and fix the initial and final learning rates to $10^{-3}$ and $10^{-6}$, respectively. We also did not use discounted reward as there was no noticeable gain, setting $\gamma = 1$. Finally, we limit the training of our methods to 200 epochs and applied early stopping using a validation set.

\subsection{Classification Results}
Table~\ref{tab:acc} shows the average classification accuracy, over $5$-fold cross-validation, of the compared methods. From the results, we can see that our proposed model is always among the top-2 in terms of performance on all tested datasets. In particular, the attention model with memory performs the best on four of the five datasets and comes in at second on the fifth dataset (NCI-33). In every single case, {\methodmem } outperforms {\method } which shows that adding an external memory to integrate information from various locations is beneficial. However, we find that {\method } still performs respectably against the compared baselines and in fact comes in second on two of the tested datasets. We also find that {\method } outperforms Agg-Attr and Agg-WL in almost every single case, which is remarkable since each agent in {\method } only has access to a portion of the graph while the latter two have access to the entire graph. In our experiments, we find that the first two baselines perform the worst, almost always performing the worst on all the datasets. The kernel-based approaches are better, with the graphlet-based approach being superior. It is able to outperform {\method } slightly. However, {\methodmem } is consistently the best performer on all the datasets that were tested.

\subsubsection{Applying Random Attention}
Our experiments show that the attention model is competitive against baselines that observe the entire graph while our model is limited to seeing a portion of the graph. To demonstrate the effectiveness of attention further, we ran another experiment where we restrict the first two baselines to the setting of \method. It is a straightforward modification since the methods also use the graph attribute vectors. However, the baselines do not have a concept of attention, so we use random attention where we sample 20 subgraphs from each graph using a random-walk of length 12. This limits the information available to the baselines to that which is available to {\method } since we fixed $M = 20$ and $T = 12$. 

Table~\ref{tab:active} shows the result of the baselines when they only observe a random portion of each graph. It is clear that the performance deteriorates for both methods, with Agg-WL showing a more marked difference in performance. This is with the exception of Agg-Attr on NCI-83. In fact, we can see that the performance of Agg-WL drops so drastically that it performs almost no better than random guessing on four of the five datasets (NCI-1, NCI-33, NCI-83, and NCI-123). This shows that attention can help us examine parts of the graph that are relevant.

\subsection{Parameter Study}
We study the effect of varying step sizes $T$ on performance of both {\method } and \methodmem. For each of the 5 datasets, we fixed all other parameters to the ones that yielded the best results and varied $T = \{1, 3, \cdots, 15, 18\}$. In both cases, accuracy increased as we increased the number of steps with $T=12$ giving fairly good performance on all datasets on both methods. Surprisingly, we found that both models already performed relatively well when $T \leq 3$, in some cases being only $5$-$6$\% worse than the best accuracy. This may be because molecular graphs are fairly small in size. We found that \methodmem, in general, benefits more from an increase in the size of $T$ which may be due to the fact that we are using weighted pooling of the history vectors so the model can support longer walks.

\section{Conclusion}
In this work, we introduced a recurrent neural network model that uses attention to process portions of a graph for classification. We also tested a variant with a more refined attention and external memory. We find that the method can outperform baselines that observe the entire graph. 

There are a lot of interesting directions for future work. We intend to study the model using more expressive node typing strategies. We would also like to experiment with an extension of the model with a more sophisticated external memory (\textit{e.g.} making memory rewritable, and using memory to condition the output of the rank network). Finally, it would be interesting to test more flexible architectures for LSTM like Tree-LSTMs that seem more natural for graphs.

\bibliography{aaai}
\bibliographystyle{aaai}
\end{document}